\documentclass[12pt]{article}

\usepackage{amsmath,amssymb}

\usepackage{helvet}

\usepackage[section]{placeins} 

\usepackage{graphicx}

\usepackage[left=0.75in,top=0.75in,right=0.75in, bottom=0.75in]{geometry}

\usepackage{cite}
\usepackage{color}
\usepackage{array}
\usepackage{setspace}
\usepackage{booktabs}
\usepackage{todonotes}
\usepackage{bm}
\usepackage{upgreek}
\usepackage{subfigure} 
\usepackage{algorithmic}
\usepackage{algorithm}
\usepackage{caption} 

\newcommand{\Ourmodel}{KG-ETM}
\newcommand{\ud}{\textrm{d}}
\newcommand{\bupbeta}{\bm{\upbeta}}
\newcommand{\bupalpha}{\bm{\upalpha}}
\newcommand{\buprho}{\bm{\uprho}}
\newcommand{\bupdelta}{\bm{\updelta}}

\usepackage{xr}
\usepackage{url}
\usepackage{multirow}
\usepackage{setspace}

\usepackage[utf8]{inputenc} 
\usepackage[T1]{fontenc}    
\usepackage{url}            
\usepackage{amsfonts}       
\usepackage{nicefrac}       
\usepackage{microtype}      

\usepackage{lineno}
\usepackage{longtable}
\usepackage{lscape}
\usepackage[final]{pdfpages}

\usepackage[document]{ragged2e}

\definecolor{DarkRed}{rgb}{0.5,0.1,0.1}

\usepackage{hyperref}

\usepackage{authblk}

\usepackage{abstract}

\usepackage{adjustbox}
\usepackage{upgreek}
\usepackage{bm}

\usepackage{multirow}
\usepackage{longtable}
\usepackage{colortbl}

\title{Modeling electronic health record data using a knowledge-graph-embedded topic model}
\author[1]{Yuesong Zou}
\author[1]{Ahmad Pesaranghader}
\author[2]{Aman Verma}
\author[2]{David Buckeridge}
\author[1,$^*$]{Yue Li}
\affil[1]{School of Computer Science, McGill University}
\affil[2]{School of Population and Global Health, McGill University}
\affil[*]{Correspondence to yueli@cs.mcgill.ca}

\date{}

\begin{document}

\maketitle

\setlength{\parindent}{1em}


\section*{\centering ABSTRACT}
The rapid growth of electronic health record (EHR) datasets opens up promising opportunities to understand human diseases in a systematic way. However, effective extraction of clinical knowledge from the EHR data has been hindered by its sparsity and noisy information. We present \Ourmodel, an end-to-end knowledge graph-based multimodal embedded topic model. \Ourmodel~distills latent disease topics from EHR data by learning the embedding from the medical knowledge graphs. We applied \Ourmodel~to a large-scale EHR dataset consisting of over 1 million patients. We evaluated its performance based on EHR reconstruction and drug imputation. \Ourmodel~demonstrated superior performance over the alternative methods on both tasks. Moreover, our model learned clinically meaningful  graph-informed embedding of the EHR codes. In additional, our model is also able to discover interpretable and accurate patient representations for patient stratification and drug recommendations. Our code is available at \href{https://anonymous.4open.science/r/Knowledge_graph-ETM-FD6F}{Anonymous GitHub}. 

\section{Introduction}
\label{sec:intro}

The rapid growth in volume and diversity of electronic health record (EHR) is prompting health informatics research to improve care for individual patients. Modern hospitals routinely generate standardized EHR observations such as International Classification of Diseases (ICD) for diagnoses, Drug Identification Number (DIN) for prescription, and Anatomical Therapeutic Chemical code (ATC) for drug ingredients. ICD is a widely used health care classification system used to classify diseases, symptoms, signs, abnormal findings, social circumstances, complaints and external causes of injury or disease. A DIN code uniquely identifies all drug products sold in a dosage form in Canada and is located on the label of prescription and over-the-counter drug products that have been evaluated and authorized for sale in Canada. ATC is a medicine classification system maintained by the World Health Organization (WHO). Each ATC code is specific to an active drug ingredient, which is indicative of patient health state. 

The rich patient EHR information enables computational phenotyping \cite{subtypingbaytas2017patient}, risk prediction \cite{riskcheng2016risk}, patient stratification \cite{stratificationlandi2020deep}, and patient similarity analysis \cite{similarityzhu2016measuring}. In this paper, we focus on computational phenotyping tasks, that is, to transform the noisy, massive electronic health record (EHR) data into meaningful medical concepts that can be further used for downstream tasks, e.g. to predict the response to drug therapy. 

Extracting meaningful medical concepts by modeling the joint distribution of the EHR data is challenging because of the large EHR feature space. Among diverse machine learning approaches, topic models provide an efficient way to exploit sparse and discrete data. They were  originally developed to discover patterns of word usages from corpora of text documents \cite{blei2003latent}. A topic model infers a set of categorical distributions called latent topics over the vocabulary and represents each document by the topic mixture memberships over those topics. Here we consider each patient's EHR history as a document and each EHR observation (e.g. ICD code) as a word within its document. Our goal then is to learn clinical meaningful disease topics and disease mixture memberships for the patients. Recently, several topic methods were developed to effectively infer latent topics from EHR data \cite{li2020mixehr,ahuja2020surelda,ahuja2021mixehr,song2021supervised}. However, these methods have limited performance in modeling some rare diseases due to insufficient EHR observations. 

In this paper, we present an end-to-end knowledge-graph embedded topic model (KG-ETM). Briefly, \Ourmodel~utilizes a graph attention network (GAT) to compute the embedding of EHR codes from a graph consisting of the taxonomies of the ICD code and ATC code as well as the connections between drugs and diseases. GAT assigns different multi-head attention in the aggregation of neighbors' information according to their embedding. This enables our model to capture the various connections between medical concepts. The resulting EHR code embedding are then used to infer a set of coherent multi-modal topics from the patient-level EHR data. Learning the graph embedding and the topic embedding are performed simultaneously in an end-to-end fashion without supervision. We experimented on a large-scale EHR dataset consisting of 1.2 million patients from Quebec, Canada. \Ourmodel~demonstrates accurate EHR reconstruction and drug imputation and learns clinically meaningful topics. Additionally, \Ourmodel~learns well-organized code embeddings because of our graph design. 

\section{Our contribution} 
\label{sec:contribution}
\begin{itemize}
    \item We proposed \Ourmodel, an end-to-end deep learning framework, which simultaneously learns the medical code embedding from a medical knowledge graph of diseases (ICD-9 code) and drugs (ATC code) via a graph attention network (GAT) \cite{velickovic2018graph} and the topic embedding from EHR data via an embedded topic model (ETM) \cite{etmdieng2019topic}.  
    \item By using a linear decoder, \Ourmodel~is able to extract meaningful and highly interpretable disease topics, which can be used in downstream tasks such as patient risk stratification. 
    \item We proposed a graph augmentation to effectively encodes hierarchical information in the knowledge graphs (i.e., a rooted tree) of medical classification systems. Besides, we combined the two knowledge graphs (of ICD-9 and ATC) via known disease-drug links (i.e. drug treatments for diseases), which allows information sharing between the two data types during the training. 
\end{itemize}

\section{Related Works}
\label{sec:related}
Recently, many automatic EHR-based phenotyping algorithms were developed using rule-based \cite{delisle2013using,mo2015desiderata,xi2015identifying} or machine learning techniques \cite{henriksson2013semantic,wu2013automated,fan2013billing,afzal2013improving,shivade2014review,lipton2015learning,alzoubi2019review}. MixEHR \cite{li2020mixehr} extended latent Dirichlet Allocation (LDA) \cite{blei2003latent} to multi-modality and accounts for missing data and non-randomly missing mechanism in the EHR data. However, MixEHR is unable to make use of knowledge graphs. In order to achieve better performance in modelling rare disease phenotypes from EHR data, several methods utilized resourceful medical domain knowledge graphs. For instance, GRAM \cite{choi2017gram} and KAME \cite{ma2018kame}  employ attention mechanism to incorporate medical knowledge into clinical modelling. GRAM considered taxonomic hierarchy as a knowledge prior and generates representation of medical concepts for a predictive task. KAME only utilized the medical knowledge related to the last visit in a recurrent neural network (RNN). RETAIN \cite{choi2016retain,kwon2018retainvis} is a two-level attention model that detects influential past visits and crucial clinical variables within those visits. DG-RNN \cite{yin2019domain}  employed an attention module that takes  long short-term memory (LSTM)'s output and models sequential medical events. To handle various healthcare tasks, TAdaNet \cite{hajij2021tda} a meta-learning model makes use of a domain-knowledge graph to provide task-specific customization. These recent models are mostly focused on supervision, and their learning algorithms require labelled data. In contrast, GETM \cite{wang2022graph} leveraged a knowledge graph by combining node2vec \cite{grover2016node2vec} with embedded topic model (ETM) \cite{dieng2020topic} in a pipeline approach. GETM is an unsupervised model that directly learn the distribvution of the EHR data using the node2vec embedding. However, because the embedding is learned separately from the ETM model, it may not always help in learning the EHR data.

\section{Methods}
\label{sec:method}

\begin{table}[t]
\caption{Notation definitions}\label{tab:notations}
    \centering
    \begin{tabular}{l|l}
    \toprule
    Notations & Descriptions\\
    \midrule
    $D$  & \# of patients in the dataset \\
    $K$     & \# of topics \\
    $c_{pn}^{(t)}$ & the $n$-th code of type $t$ of patient $p$ \\
    $V_{\text{icd}}, V_{\text{atc}}$ & size of ICD, ATC vocabulary, respectively \\
    $N_{pt}$ & \# of observed EHR codes of type $t$ for patient $p$ \\
    $\textbf{v}_p\in \mathbb{N}^{V_{\text{icd}}+V_{\text{atc}}}$ & observed code frequency for patient $p$ \\    
    $\theta_p\in S^{K-1},$  & topic mixture of patient $p$, $\sum_k\theta_{pk}=1$\\
    $\buprho^{\text{(icd)}}\in \mathbb{R}^{L \times V_{\text{icd}}}$ & KG-informed embedding of ICD codes\\
    $\buprho^{\text{(atc)}} \in \mathbb{R}^{L \times V_{\text{atc}}}$ & KG-informed embedding of ATC codes\\
    $\bupalpha^{\text{(icd)}} \in\mathbb{R}^{L\times K}$ & embedding of topics for ICD code\\
    $\bupalpha^{\text{(atc)}} \in\mathbb{R}^{L\times K}$ & embedding of topics for ATC code\\
    $\bupbeta_k^{\text{(icd)}} \in S^{V_{\text{icd}}-1}$ & $k^{th}$ ICD topic distribution, $\sum^{V_{\text{icd}}}_{v=1}\bupbeta_{vk}^{\textrm{(icd)}} = 1$ \\
    $\bupbeta_k^{\text{(atc)}} \in S^{V_{\text{atc}}-1}$ & $k^{th}$ ATC topic distribution, $\sum^{V_{\text{atc}}}_{v=1}\bupbeta_k^{\textrm{(atc)}} = 1$ \\
    \bottomrule
    \end{tabular}
\end{table}

\subsection{Notations}
We denote the number of patients, the number of topics, the size of ICD vocabulary, and the size of ATC vocabulary as $D$, $K$, $V_{\text{icd}}$, $V_{\text{atc}}$, respectively. For a patient $p$,  $c_{pn}^{(t)}$ denotes the $n$-th code of type $t\in\{\textrm{ICD}, \textrm{ATC}\}$. Another way to express the EHR history of patient $p$ is via a $(V_{\text{icd}}+V_{\text{atc}})$-dimensional frequency vector $\mathbf{v}_p$.  $\theta_p$ denotes a $K$-dimensional probabilistic topic mixture over $K$ disease topics that sum to 1. For the $k$-th topic, $\bupbeta^{(t)}_k$ denotes the code distribution of ICD/ATC code. The topic embedding is denoted by $L\times K$ matrices $\bm{\bupalpha}^{\textrm{(icd)}}, \bm{\bupalpha}^{\textrm{(atc)}}$, where $L$ is the dimension of latent embedding space. The knowledge graph (KG)-informed embedding of medical codes of type $t\in\{\textrm{icd}, \textrm{atc}\}$ is denoted by a $L\times V_{t}$ matrix $\bm{\buprho}^{(t)}$. 
Table~\ref{tab:notations} lists the key notations. 

\subsection{Generative Process}
\label{sec:generative}
\Ourmodel~assumes the following generative process (Figure \ref{fig:model}a):\\
For each patient $p \in \{1,\ldots,D\}$:
    \begin{enumerate}
        \item Draw topic mixture $\bm{\uptheta}_p \sim \mathcal{LN}(0, I)$:
            \begin{align*}
                \bm{\bupdelta}_p &\sim \mathcal{N}(\mathbf{0},\mathbf{I}),\quad 
                \mathbf{\bm{\uptheta}}_p = \frac{\exp(\bm{\bupdelta}_p)}{\sum_{k'}\exp(\bupdelta_{pk'})}
            \end{align*}
        \item For each EHR code $c_{pn}^{(t)}$, $t\in\{\textrm{icd},\textrm{atc}\}$: 
        \begin{enumerate}
            \item[] $c^{(t)}_{pn}\sim Cat(\bm{\bupbeta}^{(t)}\bm{\uptheta}_p)$. 
        \end{enumerate}
    \end{enumerate}
where $Cat$ stands for categorical distribution. The $k^{th}$ topic distribution $\bm{\bupbeta}^{(t)}_k$ is defined by the inner product of the code embedding $\bm{\buprho}^{(t)}$ and topic embedding of the $k^{th}$ topic $\bm{\bupalpha}_{\cdot k}$:
\begin{align}\label{eq:beta}
    \bm{\bupbeta}^{(t)}_k = \text{softmax}({\bm{\buprho}^{(t)}}^\intercal \bm{\bupalpha}_{\cdot k})
    = \frac{\exp({\bm{\buprho}^{(t)}}^\intercal \bm{\bupalpha}_{\cdot k})}
    {\sum_v\exp({\bm{\buprho}_{v.}^{(t)}}^\intercal \bm{\bupalpha}_{\cdot k})}
\end{align}
where ${\bm{\buprho}^{(t)}_{v.}}^\intercal$ is the $1 \times L$ row embedding of code $v$ of type $t$ and $\bm{\bupalpha}_{\cdot k}$ is the $L\times 1$ column embedding of topic $k$.

\subsection{Evidence lower bound}
\label{sec:elbo}
The marginal log-likelihood of the EHR corpus is: 
\begin{align}
    \log p(\mathbf{V} \mid \buprho,\bupalpha) 
     &= \sum_{p=1}^D \int \log p(\bm{\uptheta}_p) p(\mathbf{v}_p \mid \bm{\uptheta}_p,  \buprho,\bupalpha) \ud \theta_p\notag\\
     &= \sum_{p=1}^D \int \log p(\theta_p) \ud \bm{\uptheta}_p + \sum_{t\in\{\textrm{icd}, \textrm{atc}\}}\sum^{N_{pt}}_{n=1} \log \bupbeta_{c_{pn}^{(t)}\cdot}^{(t)} \bm{\uptheta}_p \ud \bm{\uptheta}_p \label{likelihood}
\end{align}
The marginal likelihood involves an intractable integral over the K-dimensional latent variable $\bm{\uptheta}_p$. To approximate the log-likelihood, we took an variational autoencoder (VAE) approach using a proposed Gaussian distribution $q(\bm{\uptheta}_p \mid \mathbf{v}_p, \mathbf{W})$, which is a parameterized  by a set of neural network parameters $\mathbf{W}$ \cite{kingma2013auto}. We optimize the network parameters $\mathbf{W}$ by maximizing the following evidence lower bound (ELBO):
\begin{align}
    \log p(\mathbf{V} \mid \buprho,\bupalpha) 
     & \ge  \sum_p \mathbb{E}_{q(\bm{\uptheta}_p; \textbf{v}_p, \mathbf{W})}\left[ \log p(\textbf{v}_p | \bm{\uptheta}_p, \buprho, \bupalpha)\right] \notag
     - \sum_p  \textrm{KL}\left[q(\bm{\uptheta}_p \mid \textbf{v}_p, \mathbf{W}) || p(\bm{\uptheta}_p)\right] \\
     & \equiv ELBO(\mathbf{W},\buprho,\bupalpha)
     \label{eq:elbo}
\end{align}
where the first term is the approximated log likelihood and the second term is the KL divergence between the proposed distribution and the prior for $\bm{\uptheta}_p$.

\subsection{Inferring patients' topic mixture}\label{sec:theta}

To infer $q(\bm{\uptheta}_p \mid \mathbf{v}_p, \mathbf{W})$ using VAE, we have the  following encoder architecture. Given an EHR document of two data types $\mathbf{v}_p=[\mathbf{v}_p^{\textrm{(icd)}} || \mathbf{v}_p^{\textrm{(atc)}}]$, the encoder has two input layers with rectified linear unit (ReLU) activation that separately encode $v_p^{\textrm{(icd)}}$ and $v_p^{\textrm{(atc)}}$ onto two 128-dimensional vectors $\textbf{e}_p^{\textrm{(icd)}}, \textbf{e}_p^{\textrm{(atc)}}$. We then perform a element-wise addition of the encoding vectors. The resulting 128-dimensional vectors is passed to a two fully-connected feedforward functions $\textbf{NN}_{\bm{\upmu}}$ and $\textbf{NN}_{\bm{\upsigma}}$ to generate the mean and $\log$ standard deviation of the proposed distribution $q(\bm{\uptheta}_p \mid \mathbf{v}_p, \mathbf{W})$ for patient $p$: 
\begin{align}
    \bm{\upmu}_p &= \textbf{NN}_{\bm{\upmu}}(\textbf{e}_p^{\textrm{(icd)}}+\textbf{e}_p^{\textrm{(atc)}}; \mathbf{W}_{\bm{\upmu}}),  \\
    \log\bm{\upsigma}_p &=  \textbf{NN}_{\bm{\upsigma}}(\textbf{e}_p^{\textrm{(icd)}}+\textbf{e}_p^{\textrm{(atc)}}; \mathbf{W}_{\bm{\upsigma}})
\end{align}

\subsection{Learning medical code embedding from knowledge graph}
\begin{figure*}[!ht]
  \centering
  \includegraphics[width=\textwidth]{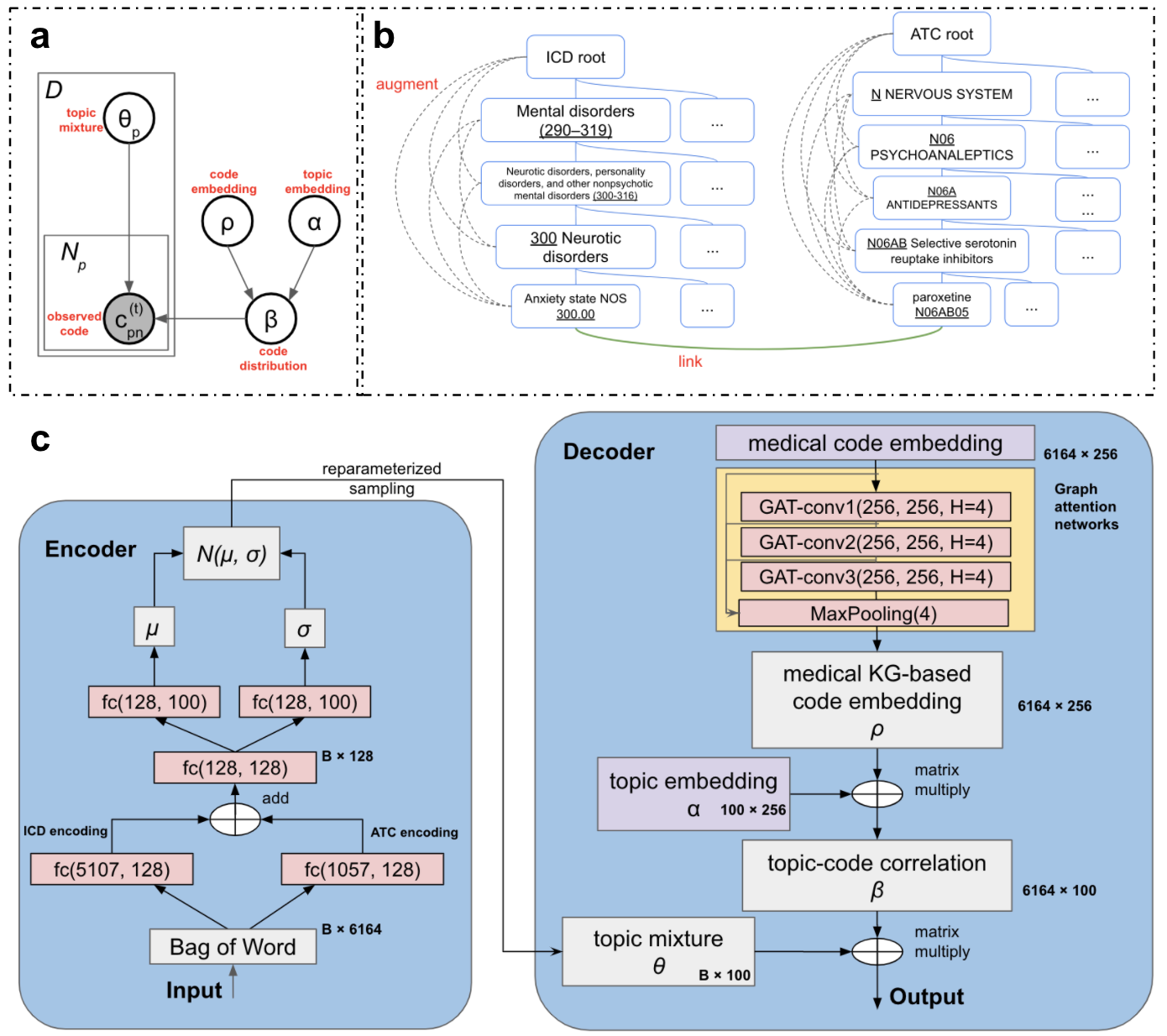}
  \caption{(a). The probabilistic graphical model view of \Ourmodel. (b).  The augmentation and merger applied on the knowledge graphs. (c). The illustration of \Ourmodel}
  \label{fig:model}
\end{figure*}

We leverage an ICD-ATC knowledge graph to learn code embedding $\buprho^{\textrm{(icd)}}, \buprho^{\textrm{(atc)}}$. As shown in Figure~\ref{fig:model}b, there are  3 types of relations in this knowledge graph: (1) ICD hierarchy (\url{https://icdlist.com/icd-9/index}) augmented by linking each pair of descendants and ancestors, (2) ATC hierarchy (\url{https://www.whocc.no/atc_ddd_index/}) augmented by linking each pair of descendants and ancestors, (3) ICD-ATC relations (\url{http://hulab.rxnfinder.org/mia/}). We extracted these relations from their corresponding websites and constructed an undirected knowledge graph $\mathcal{G}=\{\mathcal{V}, \mathcal{E}\}$, where $\mathcal{V}$ contains all of the ICD and ATC codes as the nodes and $\mathcal{E}$ contains ICD-ICD, ATC-ATC, and ICD-ATC relations as the edges. 

The resulting knowledge graph is sparsely connected because of the tree-structured of both ICD and ATC taxonomy. To improve information flow, we augmented the knowledge graph by connecting each node to all of its ancestral nodes (Figure \ref{fig:model}b).

To learn the node embedding, we used a graph attention networks (GATs) \cite{velickovic2018graph} (Figure \ref{fig:model}c). The reason why we choose GAT among numerous graph neural networks (GNNs) is that connections between medical concepts may vary (e.g. various weights of ICD-ATC relations, direction of parent-child edges) while GAT assigns multi-head self-attention to neighbors when aggregating their information.  We omit the node type notation for the embedding $\buprho$ in this subsection since the node types are ignored in GAT. We first initialized the embedding $\buprho^{(0)}$ by training a node2vec model \cite{grover2016node2vec} on the knowledge graph with embedding dimensions set to 256. We then feed the resulting embedding as the initial embedding to the multi-layer GAT network model. We compute the embedding at the $i$-th layer as:
\begin{align}
    \buprho_c^{(i)}&= \sum_{c' \in \{c\}\cup \mathcal{N}(c)} w_{cc'}^{(i)}  \textbf{W}_i  \buprho_{c'}^{(i-1)} 
\end{align}
where $\mathcal{N}(c)$ denote the neighbor nodes of node $c$ and the attention coefficients $w_{cc'}^{(i)}$ are computed as:
\begin{align}
    w_{cc'}^{(i)}= \frac{\exp(\textrm{LeakyReLU}({\textbf{a}_i}^T[ \textbf{W}_i\buprho_c^{(i)}|| \textbf{W}_i\buprho_{c'}^{(i)}])}{\sum_{j \in \{c\}\cup \mathcal{N}(c)}\exp (\textrm{LeakyReLU}({\textbf{a}_i}^T[ \textbf{W}_i\buprho_c^{(i)}|| \textbf{W}_i\buprho_{j}^{(i)}])}
\end{align}
where $\textbf{a}_i, \textbf{W}_i$ are the parameters of the $i$-th layer of the GAT network. The output of the all layers are maxpooled to a $L \times V$ embedding matrix denoted as $\buprho$. $\buprho =[\buprho^{\textrm{(icd)}}||\buprho^{\textrm{(atc)}}]$ is used as the EHR code embeddings in Eq. \ref{eq:beta}.


\subsection{Learning procedure}
In the above model, we have a set of learnable parameters including the VAE encoder network parameters $\mathbf{W}$ for $q(\bm{\uptheta}_p \mid \mathbf{v}_p, \mathbf{W})$, the GAT network parameters $\mathbf{W}_{\rho}$ for generating the code embedding $\buprho$ , and the fixed point topic embedding $\bupalpha$. To learn them, we maximize the ELBO (Eq \ref{eq:elbo}) with respect to those parameters. Specifically, we used stochastic optimization, forming noisy gradients by taking Monte Carlo approximations of the full gradient through the reparameterization trick \cite{kingma2013auto}: 
\begin{align}
    &ELBO(\mathbf{W}_{\theta}, \mathbf{W}_{\rho}, \buprho, \bupalpha) 
    \approx \sum_{p\in\mathcal{B}} \left[\log p(\textbf{v}_p | \hat{\bm{\uptheta}}_p, \buprho, \bupalpha)\right] - \sum_{p\in\mathcal{B}} \log q(\hat{\bm{\updelta}}_p \mid \textbf{v}_p, \mathbf{W}) 
    + \log p(\hat{\bm{\updelta}}_p)
\end{align}
where $
    \hat{\bm{\updelta}}_p \sim \bm{\upmu}_p + \bm{\upsigma}_p\mathcal{N}(0,I),\;
    \hat{\bm{\uptheta}}_p = \textrm{softmax}(\hat{\bm{\updelta}}_p)$
    
We used mini-batch stochastic gradient decent to update the model with each mini-batch of size $|\mathcal{B}| << D$ to handle large EHR data collection \cite{hoffman2013stochastic}. Algorithm \ref{alg:inference} summarizes the \Ourmodel~learning procedure. 
\begin{algorithm}[!h]
   \caption{Inference algorithm of 
   \Ourmodel}\label{alg:inference}
    \begin{algorithmic}
    \STATE Initialize model and variational parameters
        \FOR{epoch $i \gets 1, 2, \ldots$}
        \STATE Compute $\buprho \gets \text{GAT}(\buprho^{(\textrm{init})}; \textbf{a}_{\textrm{GAT}}, \textbf{W}_{\textrm{GAT}})$
        \FOR{type $t \gets \text{icd}, \text{atc}$}
            \STATE {Compute $\bupbeta_k^{(t)} \gets \textrm{softmax}({\bm{\uprho}^{(t)}}^\intercal\bm{\upalpha}_k^{(t)})$ for each topic $k$ }
            \ENDFOR 
            \STATE Choose a minibatch $\mathcal{B}$ of patients
           \FOR {each patient $p$ in $\mathcal{B}$} 
                \STATE {Get normalized bag-of-word representation  $\mathbf{v}_p'$}
                \STATE Compute  $\mathbf{e}^{(\text{icd})}, \mathbf{e}^{(\text{atc})} \gets \text{FC}_{\textrm{in}} (\mathbf{v}_p'; \textbf{W}_{\textrm{in}})$
                \STATE {Compute $\bm{\upmu}_p, \log \bm{\upsigma}_p  \gets \textbf{NN}_{\bm{\upmu},\bm{\upsigma}}(\mathbf{e}_p^{\textrm{(icd)}}+ \mathbf{e}_p^{\textrm{(atc)}}; \textbf{W}_{\bm{\upmu}}, \textbf{W}_{\bm{\upsigma}})$}
                \STATE {Sample $\bm{\uptheta_p} \sim \mathcal{LN}(\bm{\upmu}_p, \bm{\upsigma}_p)$}
                \STATE {Compute  $p(\mathbf{v}_p | \bm{\uptheta}_p,\bupbeta) \gets \mathbf{v}_p^\intercal \log \bupbeta \bm{\uptheta}_p$}
            \ENDFOR

\STATE Estimate the ELBO and its gradient (backprop.)
\STATE Update GNN parameters $\textbf{W}_{\buprho}$  (i.e. $\textbf{a}_{\textrm{GAT}}, \textbf{W}_{\textrm{GAT}}$) 
\STATE Update topic embedding $\bupalpha$
\STATE Update variational parameters $\textbf{W}$ (i.e. $\textbf{W}_{\textrm{in}}, \textbf{W}_{\bm{\upmu}}, \textbf{W}_{\bm{\upsigma}}$)
        \ENDFOR
\end{algorithmic}
\end{algorithm}

\subsection{Implementation details}
We used Adam optimizer to train \Ourmodel. The learning rate was set as 0.01. We use L2 regularization on the variational parameters. The weight
decay parameter is $1.2 \times 10^{-6}$. The minibatch size is 512. We train the model for 10 epochs. 
The embedding size of topic and code embedding was set to 256. The embedding size in the inference encoder was set to 128. Empirically we found that the number of GAT layers being 3 and number of heads being 4 is the best.


\subsection{Data processing}
To evaluate our model, we used a real-world large EHR database called Population Health Record (PopHR), which was originally created for monitoring population health from multiple distributed sources \cite{PopHR, PopHR_Usage}. PopHR contains longitudinal administrative data of 1.2 million patients with up to 20-year follow-up. For each patient, we collapsed the time series data to obtain the frequency of distinct EHR codes observed over time (i.e., $\mathbf{v}_p$). We treated the frequency as an EHR document. We used two types of EHR data: (1) 5107 unique ICD-9 codes; (2) over 10,000 DIN codes. Since different DIN codes may indicate the same ingredient(s) of different strength(s), we converted the DIN code to 1057 ATC code according to their ingredient(s). 

\subsection{Evaluation metrics} 
\subsubsection{Reconstruction}
We conducted a document completion task and calculate log likelihood as the metric of predictive capacity. We split the PopHR dataset by a train-valid-test ratio $6:3:1$. We randomly divided each test EHR document into two bags of codes of equal sizes. We used the first half to predict the expected topic mixture of the test patient ($\bar{\bm{\uptheta}}_p = \textrm{softmax}(\bm{\upmu}_p$)). We then evaluated the predicted log likelihood of the second half.  

\subsubsection{Topic quality}
Since the interpretation of the topics learned by the model is also crucial, We measured the topic quality score \cite{etmdieng2019topic}, which is the product of topic coherence and topic diversity. Topic coherence  \cite{lau2014machine} measures the observed co-occurrence rate of the top codes in the same topics. It is defined as the average pointwise mutual information of two codes drawn randomly from the same document:
\begin{align}
\mathrm{TC}=\frac{1}{K} \sum_{k=1}^{K} \frac{2}{s(s-1)} \sum_{1\leq i\leq j\leq s} \frac{\log \frac{P(w_i^{(k)}, w_j^{(k)})}{P(w_i^{(k)}) P(w_j^{(k)}}}{-\log P(w_i^{(k)},w_j^{(k)})}, 
\end{align}
where $\{w_1^{(k)},\ldots,w_s^{(k)}\}$ denotes the top-$s$ codes with the highest probability in topic $k$, $P(w_i^{(k)}, w_j^{(k)} )$ is the probability of words
$w_i^{(k)}$ and $w_i^{(k)}$ co-occurring in an EHR document and  $P(w_i^{(k)})$ is the marginal probability of code $w_i^{(k)}$. Topic diversity \cite{etmdieng2019topic} measures the variety of topics since similar topics bring redundancy. It is defined as the percentage of unique codes in the top-$r$ codes across all topics: 
\begin{align}
    \mathrm{TD} &= \frac{1}{Kr} \mathrm{unique}\left(\bigcup_{k=1}^{K} \bigcup_{i=1}^r  \{w_i^{(k)}\}\right). 
\end{align}
where $\mathrm{unique}(\cdot)$ is the function to count the number of unique elements in a set. Topic quality (TQ) is defined as TC $\times$ TD. We measure TQ for ICD codes and ATC codes separately and then compute the average of them. We set $s=3, r=3$ for the calculation of TC and TD, respectively. 

\subsubsection{Drug imputation task} 
We sought to impute ATC codes based only on ICD codes. Specifically, we first inferred $\hat{\bm{\uptheta}}_p$ from input EHR of patient $p$ using only the ICD codes. We then inferred the expectation of each ACT code $\hat{c}_{pv}^{\textrm{(atc)}} = \bm{\upbeta}_v^{(\textrm{atc})}\hat{\bm{\uptheta}}_p$. 

We evaluated the models by patient-wise accuracy and drug-wise accuracy. For patient-wise accuracy, we compared the precision, recall, and F1-score of the top-5 predictions averaged over all patients (prec@5, recall@5, F1-score@5). In both training and test datasets, patients with less than 5 ATC codes were filtered out.

To measure the influence of the knowledge graph on different nodes, we then calculated the drug-wise accuracy. All the ATC codes were sorted by their frequencies and then binned into five frequency quintiles
, where 0-20\% contains the rarest ATC codes and 80-100\% contains the most frequently observed ATC codes. We then computed the recall on each ATC code and took the average (weighted by frequency) of the codes in each bin.

\subsection{Baselines}
We compared the performance of \Ourmodel~ with two embedded topic modeling approaches: 
\begin{itemize}
    \item ETM \cite{dieng2020topic} is a topic model that introduces feature embedding of words and topics. We consider it as a baseline because it has a similar generative process as \Ourmodel~but no utilization of knowledge graph.  
    \item GETM \cite{wang2022graph} is a knowledge-graph-based topic model that leverages medical taxonomy hierarchies by initializing word embedding as the output of node2vec. We considered GETM as baseline because it harnesses medical knowledge graphs but not in an end-to-end manner. 
    \end{itemize}
We set the number of topic $K$ as 100 and the number of embedding dimensions as 256 for both baselines and our approach. The number of layers of inference networks was set to 3 for ETM and GETM to fairly compared with ours.   

For drug imputation, we also evaluated two traditional approaches:
\begin{itemize}
    \item Frequency based model: we counted the occurrence of all ATC codes in the training data, and then imputed the most frequent codes for the test patients. 
    \item K nearest neighbors: for each patient in test set, we found K nearest neighbors according to its frequency vector $\textbf{v}_p$. We then averaged the ATC codes of the nearest neighbors as the ATC predictions for the test patients. We selected the optimal number of neighbors $K \in \{100, 200, 500, 1000, 5000\}$ and the best distance metrics $\in \{manhattan,  minkowski\}$ using the validation set. 

\end{itemize}

\subsection{Ablation study}
An ablation study was conducted to evaluate the three key features of \Ourmodel:
\begin{enumerate}
    \item \textbf{initialization of code embedding:} when this procedure is discarded, we randomly initialized the embedding for GAT rather than pre-trained them by node2vec. 
    \item \textbf{augmentation of knowledge graph:} when this procedure is discarded, we did not  connect each node with all of its ancestors.
    \item \textbf{graph attention networks:} when this module is discarded, we fixed the code embedding that generated by node2vec. In other word, it is equivalent to GETM with the augmented knowledge graph. 
\end{enumerate}

\section{Results}\label{sec:results}

\subsection{Reconstruction and topic quality} \label{sec:recon}
As shown in Table~\ref{tab:recon}, \Ourmodel~performed the best on both likelihood and topic quality. ETM performed worst under every metric, because of the lack of external knowledge graph. 
Compared to \Ourmodel, GETM achieved higher TD but lower TC, which means that the distribution over codes are more diverse among topics. In GETM, the code embedding was learned only from the knowledge graph and then fixed during the ETM training on the EHR dataset. In contrast, \Ourmodel~utilized a GAT to refine the code embedding, whose parameters were trained on the EHR dataset. This led to higher TC and higher overall TQ compared to GETM. 

Table~\ref{tab:abl} summarizes the results of the ablation study. All of the three features have notable impacts on the prediction performance and topic quality. Considering log-likelihood, the graph augmentation benefited the predictive power of our model the most, the GAT module that enables the end-to-end training manner came second. Considering TQ, pre-training code embedding took the most crucial role. Compared with GETM, we found that GETM with graph augmentation achieved better precision but worse TQ. It possibly due to the fact that the connection between medical concepts are not the same. This founding highlights the importance of using the GAT.

\begin{table}[!ht]
    \centering
    \caption{Reconstruction loss and topic quality}
    \label{tab:recon}
\begin{tabular}{l|c|c|c|c|c}
     \toprule
      Model & Recon. &  \multicolumn{3}{c}{Topic Quality [ICD,ATC]}   \\
     \hline 
     & NLL. & topic coherence & topic diversity & topic quality & TQ(ave.) \\
     \midrule
      ETM \cite{etmdieng2019topic} &198.26 & 0.113, 0.233 & 0.373, 0.423& 0.0421, 0.0986 & 0.0704 \\
       GETM \cite{wang2022graph}  & 184.32 & 0.167, 0.271 & \textbf{0.86}, \textbf{0.83} &  \textbf{0.1436}, 0.2249 & 0.1843  \\
      \hline 
       \Ourmodel~(proposed) & \textbf{172.69} & \textbf{0.18}, \textbf{0.314} & 0.76, 0.787 & 0.1368, \textbf{0.2471} & \textbf{0.1920}\\
     \bottomrule
\end{tabular}
\end{table}

    \begin{table}[!ht]
    \centering
        \caption{Ablation study}
   \label{tab:abl}
\begin{tabular}{l|c|c|c|c|c}
     \toprule
      Model &  Recon. &  \multicolumn{3}{c}{Topic Quality [ICD,ATC]}   \\
     \hline 
     & NLL. & topic coherence & topic diversity & topic quality & TQ(ave.) \\
     \midrule 
       \Ourmodel & \textbf{172.69} & \textbf{0.18}, \textbf{0.314} & 0.76, 0.787 & \textbf{0.1368}, \textbf{0.2471} & \textbf{0.1920}\\
     \midrule
     \Ourmodel~ (w/o init.) & 179.59 & 0.139, 0.193 & 0.573, 0.447 & 0.0796, 0.0863 & 0.0830 \\ 
     \Ourmodel~ (w/o aug.) & 181.63 & 0.162, 0.282 & 0.733, 0.75 & 0.1187, 0.2115 & 0.1651 \\
      GETM (w/ aug.) & 180.44 & 0.161, 0.282 & \textbf{0.783}, \textbf{0.807} &0.1261, 0.2276 &0.1768 \\
     \bottomrule
\end{tabular}
\end{table}

\subsection{Drug Imputation Task}
\label{sec:drug_impu}


\begin{table}[!ht]
\centering
\caption{Patient-wise imputation measurement}
 \label{tab:drug_impu}
  \begin{tabular}{l|c|c|c}
     \toprule
      Model 
     & prec@5 & recall@5 & f1-score@5 \\
     \midrule
     frequency-based model & 0.1049 & 0.0432 & 0.0577 \\
     K nearest neighbor  & 0.1606 & 0.0713 & 0.0930  \\
     \hline 
     ETM  & 0.1823 &0.0833 & 0.1075 \\
     GETM & 0.2378 & 0.1101 & 0.1418\\
     \textbf{\Ourmodel}  &\textbf{ 0.2600} & \textbf{0.1225} & \textbf{0.1569}\\
     \bottomrule
\end{tabular}
\end{table}
\begin{table}[!ht]
\centering
\caption{Drug-wise imputation measurement}
\label{tab:freq}
\begin{tabular}{l|c|c|c|c|c}
\toprule
Model& \multicolumn{5}{c}{Percentile of rareness} \\
\hline
& 20-40 & 40-60 & 60-80 & 80-100 & ave.  \\
\midrule
ETM  &  0.0039& 0.0188& 0.0479& 0.3847 & 0.3058 \\
GETM  & 0.0213 & 0.0542 & 0.0934 & 0.4352 & 0.3597 \\
\Ourmodel & \textbf{0.0345}& \textbf{0.0841}& \textbf{0.1239}& \textbf{0.4583} & \textbf{0.3815} \\
\bottomrule
\end{tabular}
\end{table}

Table~\ref{tab:drug_impu} shows the result of patient-wise imputation performance. \Ourmodel~achieved the highest score on all metrics. 
Table~\ref{tab:freq} shows the result of drug-wise accuracy. \Ourmodel~outperformed all of the baselines. Compared to ETM, \Ourmodel's accuracy@30 is 9 times better on 20-40\%, 5 times better on 40-60\%, 3 times better on 60-80\%, and 0.25 higher on 80-100\%. This indicates that the knowledge graph embedding learned by the GAT conferred larger benefits to predicting rare ATC codes.

\begin{figure}[t]
  \includegraphics[width=\linewidth]{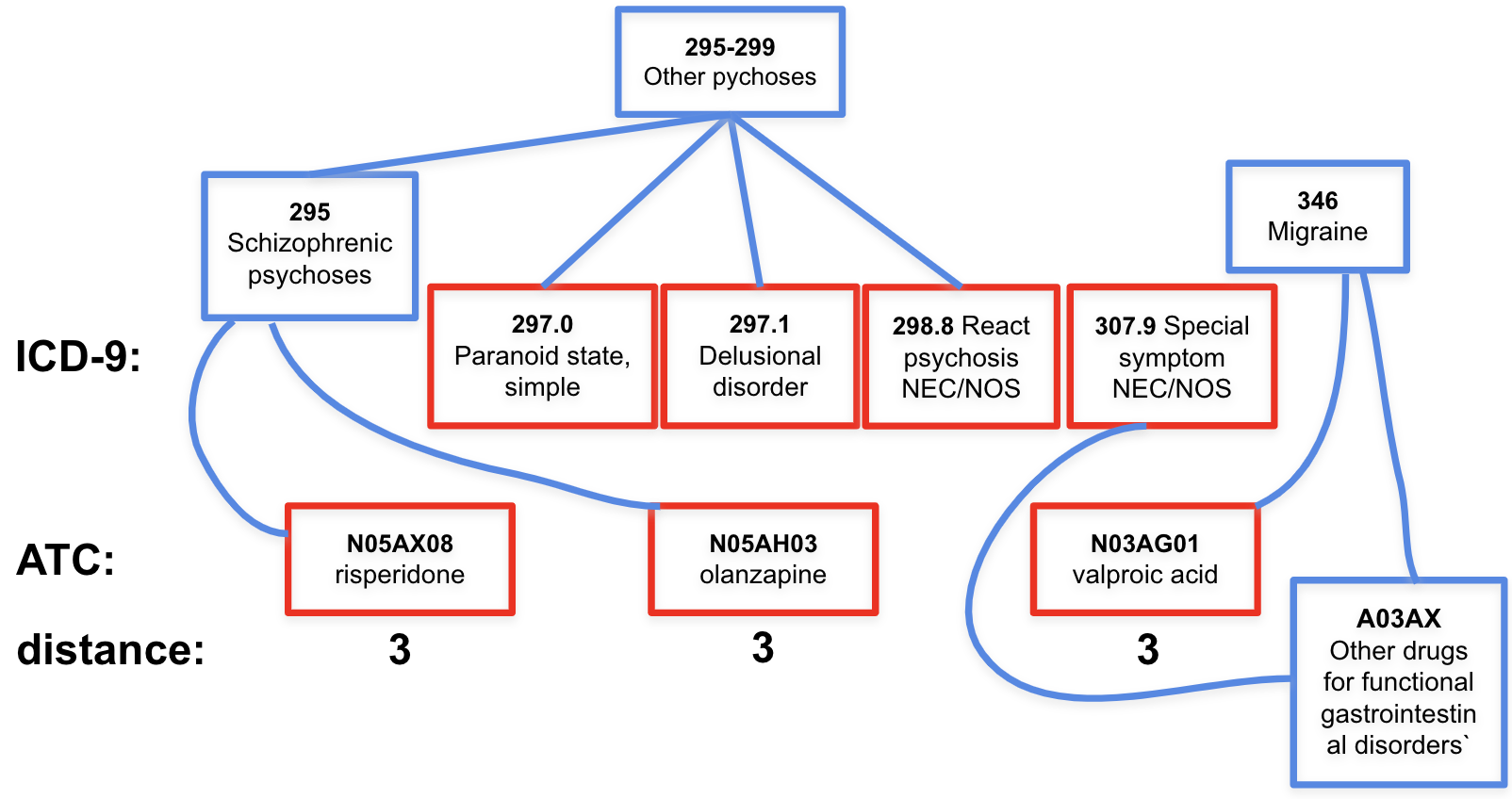}
  \caption{The distance metric example of a patient in our data. Codes in red frames are observed ICDs and imputed ATCs. The three imputed ATCs are of the same distance to observed ICDs, while their shortest paths may vary. }\label{fig:patient_a}
\end{figure}

We then conducted a case study to further ascertain our drug imputation results. For each patient, we measured the distance of each imputed ATC code from the observed ICD codes in the unaugmented knowledge graph. We collapsed the last classification level of ICD and ATC for easier analysis, while preserving sufficient granularity. Figure~\ref{fig:patient_a} illustrates an example of the distance metric. ICD codes \textit{297.0}, \textit{297.1}, \textit{298.8}, \textit{307.9} are part of observed ICD codes, and \textit{N05AX08}, \textit{N05AH03}, \textit{N03AG01} are the top 3 imputed ATC codes. Their distance to observed ICD codes are all 3. The minimal paths of the first two nodes is \textit{N05AX08}/\textit{N05AH03} $\rightarrow$ \textit{295} $\rightarrow$ \textit{295-299} $\rightarrow$ \textit{297.0}. The minimal path of \textit{N03AG01} is \textit{N03AG01} $\rightarrow$ \textit{346} $\rightarrow$ \textit{A03AX} $\rightarrow$ \textit{307.9}. 

\begin{figure*}[t]
 \includegraphics[width=\textwidth]{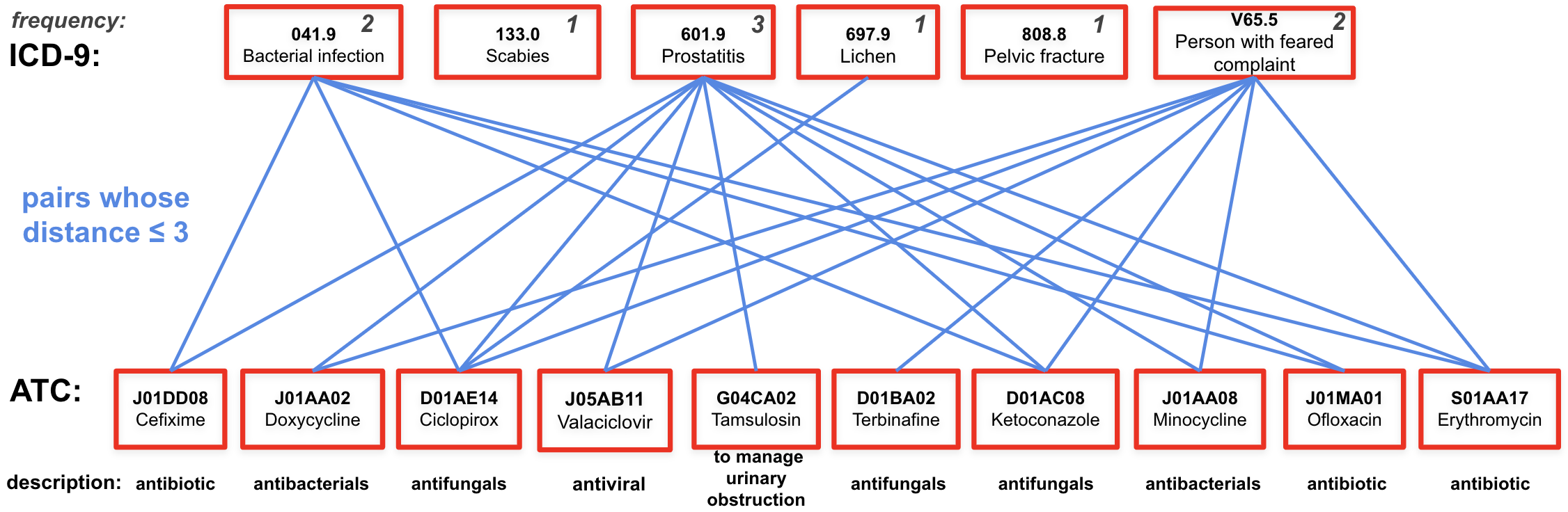}
\caption{The distance of imputed ATCs of patient e. ICD-ATC Pairs whose distances are no more than 3 are linked. We observed that the imputed ATC codes are closely connected to observed ICD codes. The within-patient frequency for each ICD code is annotated. \Ourmodel~ pays more attention to ICD codes with more occurence}\label{fig:patiente}
\end{figure*}

\begin{figure*}[!ht]
  \includegraphics[width=\linewidth]{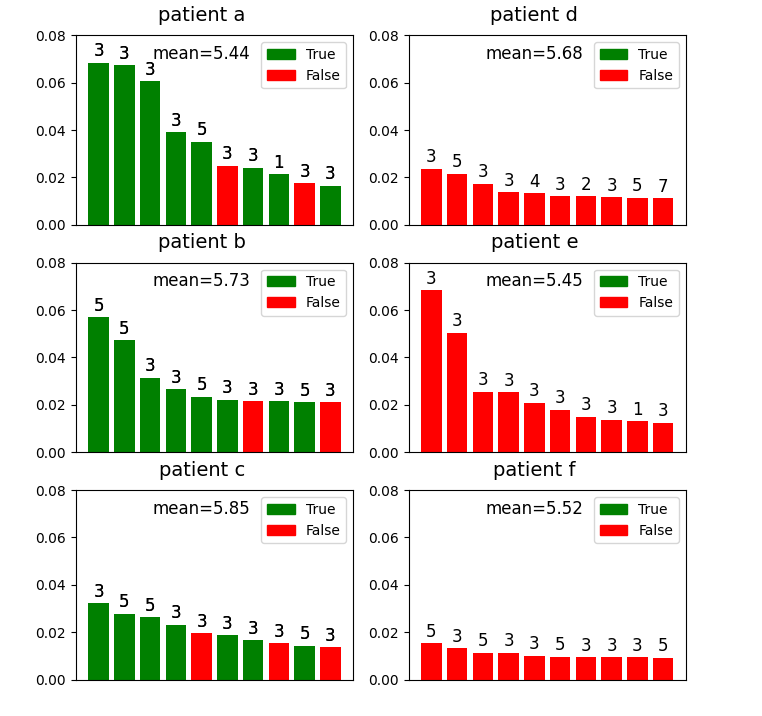}
        \caption{The distance from imputed ATCs to observed ICD. Each plot displays top 10 imputations of a patient. The height and color of a bar indicates the imputed probability and whether it is correct. Annotated above each bar is the distance from each imputed ATC to the observed ICD. } 
       \label{fig:dis-icd}
 \end{figure*}

\begin{figure*}[!ht]
   \centering
        \includegraphics[width=\linewidth]{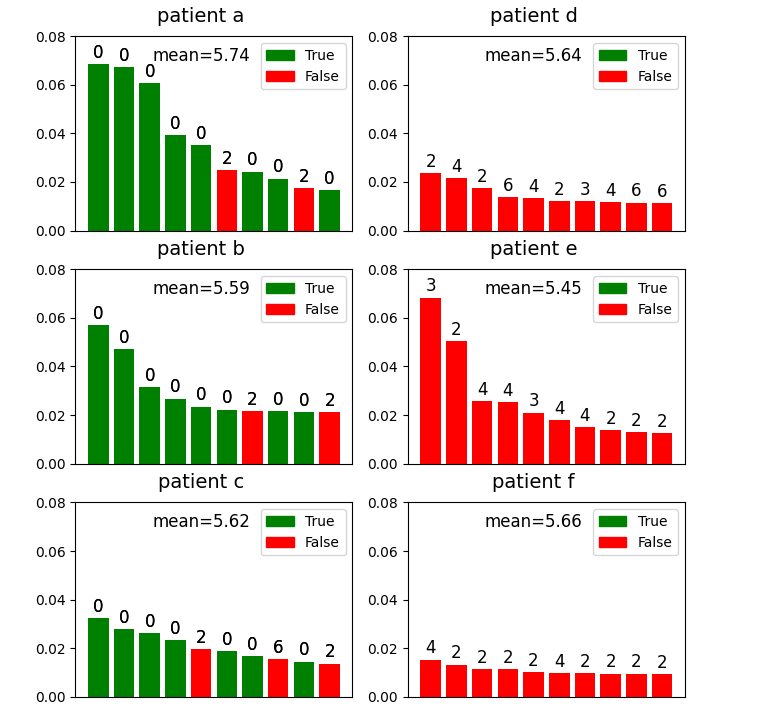}
         \label{fig:dis-atc}
         \caption{The distance from imputed ATCs to true ATC codes. Each plot displays top 10 imputations of a patient. The height and color of a bar indicates the imputed probability and whether it is correct. Annotated above each bar is the distance from each imputed ATC to the observed ATC  codes.}
\end{figure*}

Figure~\ref{fig:dis-icd} and \ref{fig:dis-atc} displays the distance of imputed top 10 ATC codes in most three accurate patients (a, b, c) and the most three inaccurate patients (d, e, f). We computed the average distance to the observed ICD codes for comparison. All imputed codes are of a distance lower than average, regardless of the accuracy. That means,  the recommended ATC codes for the inaccurate patients, though do not occur on prescription, are  highly related to observed ICD codes. Figure~\ref{fig:patiente} displays the observe ICD codes and top 10 recommended ATC codes of patient e. The recommended drugs are close to observed diseases and indeed commonly used to treat them according to the descriptions. Moreover, our model  paid more attention to diseases with higher dominance.

\subsection{Qualitative Analysis}
\subsubsection{Topic Analysis}

\begin{figure*}[!ht]
  \centering
  \includegraphics[width=\textwidth]{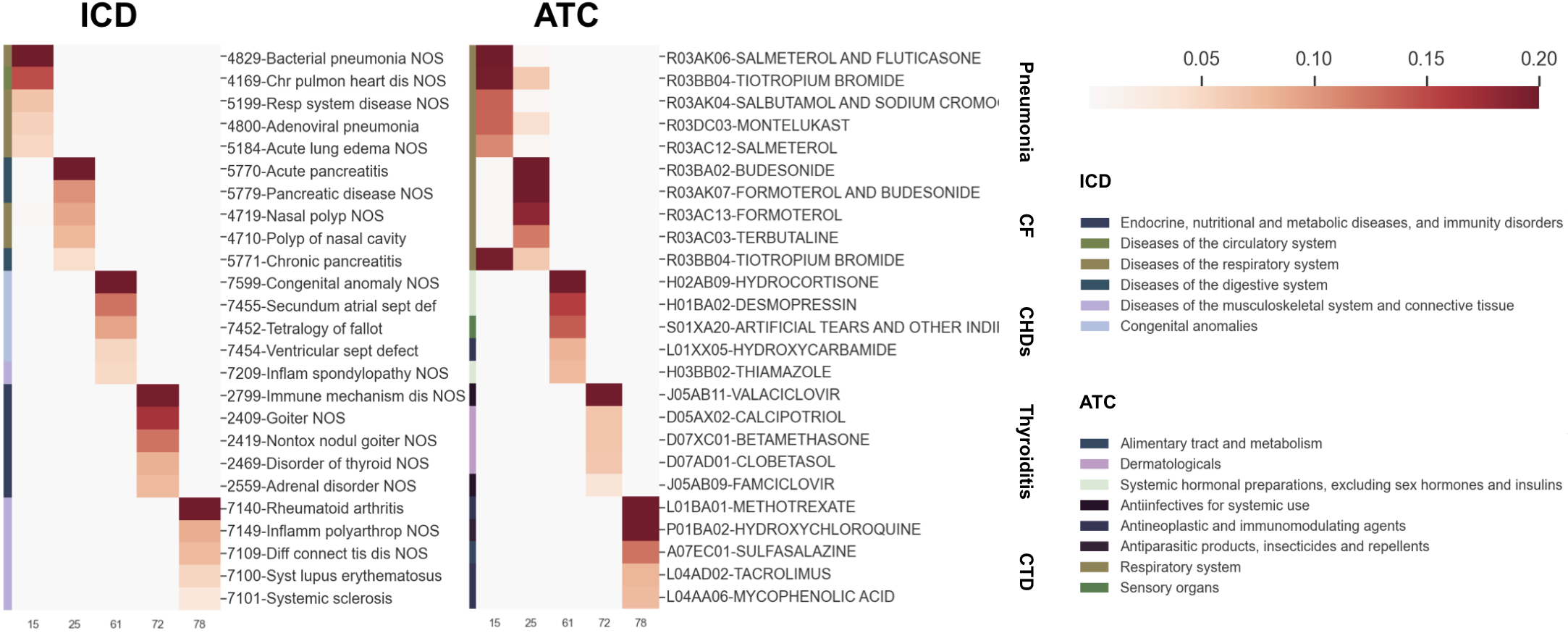}
  \caption{Heatmap of top EHR codes of 5 select topics for a diverse set of conditions. The top 5 ICD and ATC codes were displayed for the same topics in the two separate heatmaps. The heatmap intensity is proportional to the probabilities of each code under the topic.}
  \label{fig:topic_heatmap}
\end{figure*}

Figure~\ref{fig:topic_heatmap} displays 5 topics and their top 5 ICD and ATC codes. The 5 topics 15, 25, 61, 72, 78 are according with pneumonia, cystic fibrosis (CF), congenital heart defects (CHDs), thyroiditis, connective tissue diseases (CTD). Figure~\ref{fig:topic_heatmap} exhibits high intra-topic coherence and inter-topic diversity. Noticeably, CF also causes severe damage to the lung and respiratory system. Hence there are overlap between key codes of ATC codes in topic 15 (CF) and 25 (pneumonia). The color bar on the left indicates the 1st-level category each code belongs to. In Many topics codes are from the same categories. In other word, they are in the same subtree of ICD or ATC hierarchy. But there are still topics whose key codes are not in the same categories while this result makes sense according to healthcare knowledge, e.g. topic 25 cystic fibrosis triggers both lung diseases and respiratory diseases. Based on this, we claim that the knowledge graph benefits topic learning but doesn't overpower the topic model. 

\subsubsection{Medical Code Embedding}
\begin{figure*}[!h]
  \centering
  \includegraphics[width=\textwidth]{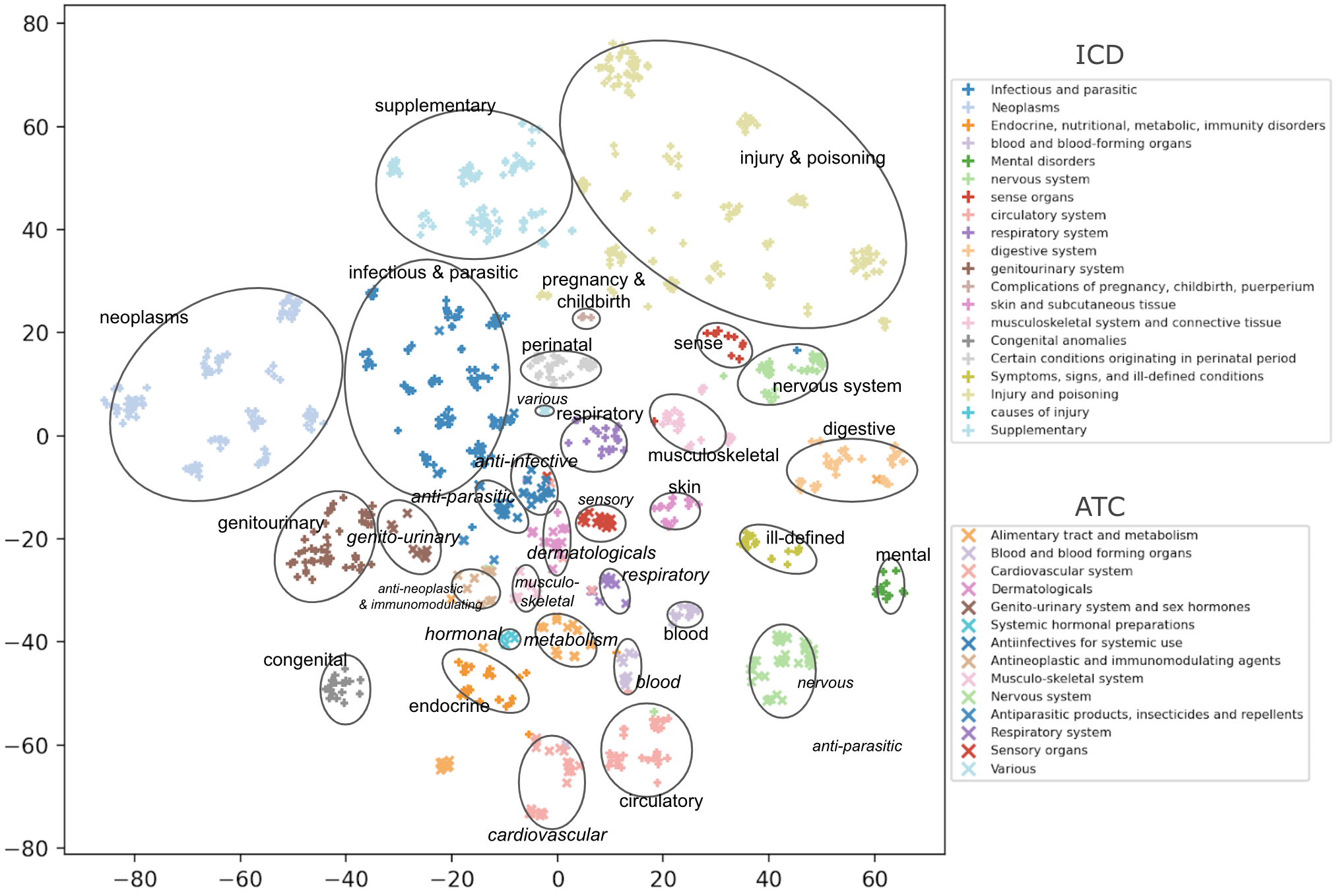}
  \caption{t-SNE plot of code embedding $\buprho$. tSNE is utilized to reduce the dimensions of code embeddings from $L$ to 2. For each point, shape ("$+ / \times$") indicates ICD / ATC, color indicates its category according to the legend. (Partially) aligned ICD and ATC categories are assigned identical (or similar) colors. Within ICD / ATC vocabularies, nodes of the same categories (colors) are grouped together. Each group was circled and labeled, and we labeled ICD and ATC groups by normal and italic fonts, respectively. Between ICD and ATC, the related categories (groups of similar colors)  are close.}
  \label{fig:code_tsne}
\end{figure*}

Figure~\ref{fig:code_tsne} displays the t-SNE plot of code embedding $\buprho^{\text{(icd)}}, \buprho^{\text{(atc)}}$. We utilized T-distributed Stochastic Neighbourhood Embedding (tSNE) to reduce the
dimensionality of code embedding from $L$ to 2. For each point, its shape ``$+ / \times$'' and color  indicate that it is an ICD / ATC code and belongs to which category. Categories are the 1st classifiction level in ICD and ATC codes. We aligned corresponding categories between ICD and ATC by assigning identical or similar colors to them, e.g.  \textit{ICD: 13-skin and subcutaneous tissue}  and  \textit{ATC: 4-dermatologicals}  are of the same color since they are perfectly aligned,  \textit{ICD: 3-endocrine, nutritional and metabolic diseases, and immunity disorders}  and  \textit{ATC: 1-Alimentary tract and metabolism}  have similar color since they are partially aligned. In the plot, codes with the same ICD / ATC categories are grouped together pretty well. We circled code for all categories and labeled  ICD groups in normal fonts, ATC in italic. The codes in related categories are close to each other, e.g.  (\textit{ICD: 1-infectious and parasitic}, \textit{ATC: 7-antiinfectives for systemic use}  , \textit{ATC: 11-antiparasitic products, insecticides and repelients}),  (\textit{ICD: 3-endocrine, nutritional, metabolic, immunity disorders} , \textit{ATC: 1-alimentary tract and metabolism}),  (\textit{ICD: 8-circulatory} ,  \textit{ATC: 3-cardiovascular system}), etc. Interestingly, the medical code embedding shows relevance between some pairs of different categories, e.g. (\textit{ATC: 4-dermatologicals}, \textit{ATC: 12-respiratory system}).

\section{Discussion}
\label{sec:discuss}
In this study, we present an end-to-end graph-embeded topic model that: (1) learns interpretable topic and code embeddings in the same embedding space; (2) is able to handle multi-modal data; (3) leverages a medical knowledge graph to improve performance quantitatively and qualitatively. We studied the performance of \Ourmodel~against several topic models on EHR reconstruction task and drug imputation tasks. \Ourmodel~outperformed the alternative methods on both tasks. Qualitative analysis further illustrated that \Ourmodel~learned  coherent topics in healthcare as well as meaningful latent embedding of medical codes. As a future work, we will construct larger graphs with richer knowledge that comprise not only ICD codes and ATC codes but also other codes available from Universal Medical Language System. Besides, we will extend GAT to multi-relational graphs to account for heterogeneous graphs.



\clearpage
\section{Figures}



\section{Tables}

\bibliographystyle{unsrt}
\bibliography{main}

\end{document}